\documentclass[letterpaper, 10 pt, conference]{ieeeconf}  %

\IEEEoverridecommandlockouts                              %

\usepackage{graphicx}
\usepackage{color}
\usepackage{multirow}
\usepackage{float}
\usepackage{tikz}
\usetikzlibrary{patterns}
\usepackage{booktabs}
\usepackage{hyperref}

\usepackage{cleveref}
\usepackage{pdfpages}

\usepackage{pgfplots}
\usepgfplotslibrary{colorbrewer}
\pgfplotsset{compat = 1.15, cycle list/Set1-8} 
\usetikzlibrary{pgfplots.statistics, pgfplots.colorbrewer} 
\usepackage{pgfplotstable}
\usepackage{filecontents}

\definecolor{cred}{RGB}{250, 104,95}
\definecolor{corange}{RGB}{255, 134,13}
\definecolor{cblue}{RGB}{127, 158,202}
\definecolor{cgreen}{RGB}{142, 187,118}

\newcommand{\link}[1]{{\color{blue}\href{#1}{#1}}}

\title{\LARGE \bf
Radiance Fields for Robotic Teleoperation
}

\author{Maximum Wilder-Smith, Vaishakh Patil, Marco Hutter%
\thanks{Authors are with the Robotic Systems
Lab, ETH Zurich. Email:  {\tt\small \{mwilder, patilv,
mahutter\}@ethz.ch}}%
\thanks{This work was funded by NCCR Automation, and Swiss Federal Railways (SBB) via ETH Mobility Initiative.}
}

\begin{document}

\maketitle
\thispagestyle{empty}
\pagestyle{empty}

\begin{abstract}

Radiance field methods such as Neural Radiance Fields (NeRFs) or 3D Gaussian Splatting (3DGS), have revolutionized graphics and novel view synthesis.
Their ability to synthesize new viewpoints with photo-realistic quality, as well as capture complex volumetric and specular scenes, makes them an ideal visualization for robotic teleoperation setups. Direct camera teleoperation provides high-fidelity operation at the cost of maneuverability, while reconstruction-based approaches offer controllable scenes with lower fidelity. 
With this in mind, we propose replacing the traditional reconstruction-visualization components of the robotic teleoperation pipeline with online Radiance Fields, offering highly maneuverable scenes with photorealistic quality. As such, there are three main contributions to state of the art: (1) online training of Radiance Fields using live data from multiple cameras, (2) support for a variety of radiance methods including NeRF and 3DGS, (3) visualization suite for these methods including a virtual reality scene. To enable seamless integration with existing setups, these components were tested with multiple robots in multiple configurations and were displayed using traditional tools as well as the VR headset. The results across methods and robots were compared quantitatively to a baseline of mesh reconstruction, and a user study was conducted to compare the different visualization methods. The code and additional samples are available at \link{https://rffr.leggedrobotics.com/works/teleoperation/}.
\end{abstract}

\section{Introduction}

Robotics is rapidly expanding its role in our everyday activities, which needs fast, repetitive, and precise actions. Robots are becoming highly autonomous, from vacuum cleaners to everyday commute vehicles. However, autonomous robots are still limited by perception, planning, and control capabilities for different tasks. Here, they still need human intervention to be successfully executed. The process of controlling robots remotely is known as robotic teleoperation. As robots become increasingly integrated into complex environments, there is a growing need for teleoperation systems that offer high-fidelity reconstruction and intuitive user interfaces. For the robots to be teleoperated by humans, the system must create an immersive environment that can provide situational awareness to the operator. Traditional teleoperation pipelines capture sensor data from robots and display it to an operator. Direct camera streams offer high fidelity at the cost of maneuverability, while reconstructed environments are easy to control but often struggle to accurately capture scene geometry and texture, especially in volumetric or specular environments. The reconstruction quality is often limited by high sensor noise and temporal inconsistencies~\cite{vrteleop}. Moreover, the user interfaces may lack immersion and intuitive interaction, limiting the operator's ability to perform tasks effectively. The proposed pipeline addresses these challenges by leveraging advancements in Radiance Fields and immersive visualization technologies to bring high-fidelity, maneuverable scenes.

\begin{figure}[t]
    \centering
    \includegraphics[width=\linewidth]{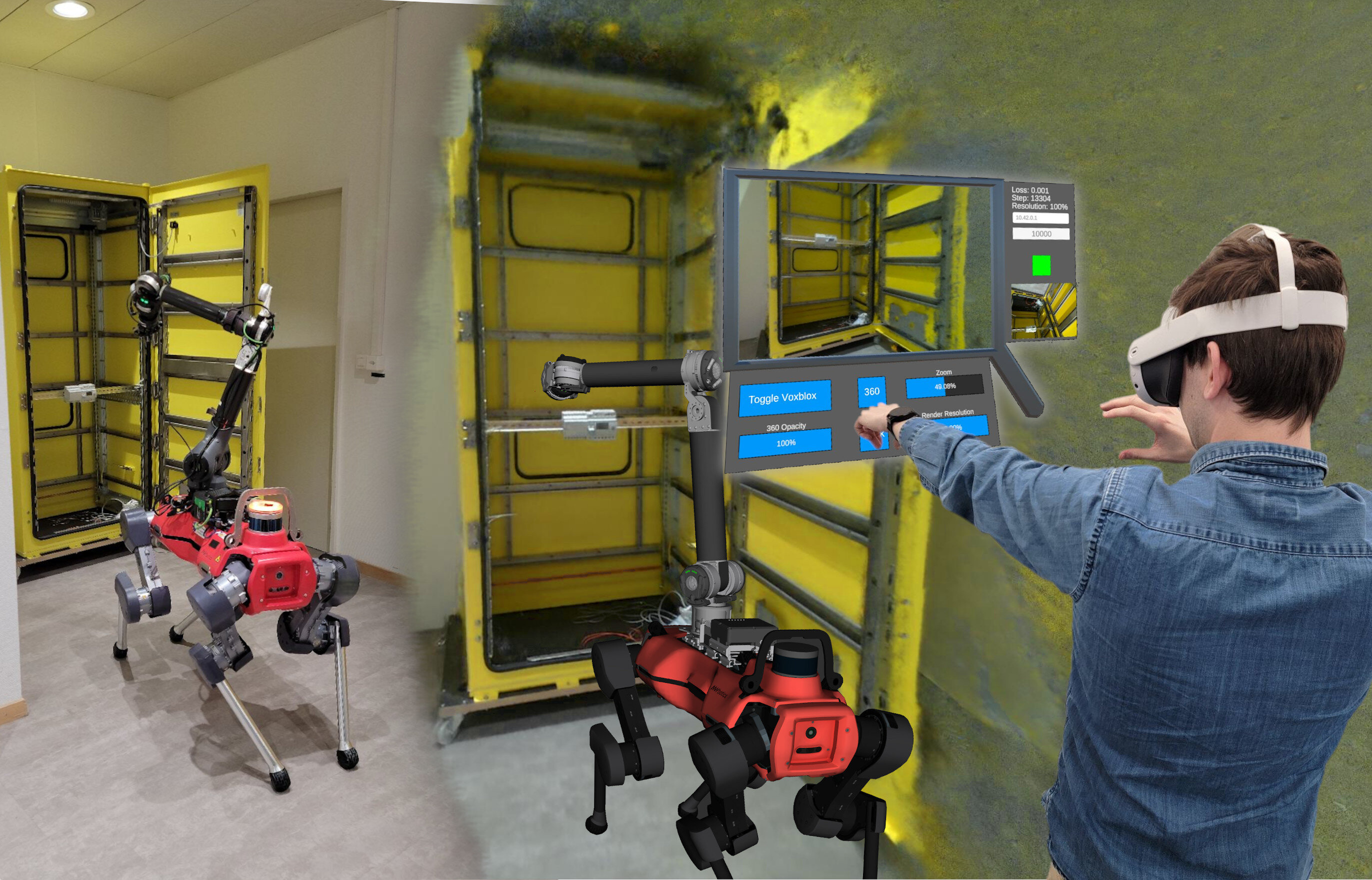}
    
    \caption{Teleoperator controlling a robot using a VR interface from inside a reconstructed Radiance Field. Reconstructions are created online from any robot based on its sensor configuration. Neural Radiance Field or Gaussian Splatting renders can be displayed in an immersive 360$^{\circ}$ render or on a handheld viewer. These renders are displayed alongside the robot and its sensor data, such as camera feeds and LiDAR.}\vspace{-15px}
    \label{fig:showcase}
\end{figure}

\begin{figure*}[h]
    \centering
    \includegraphics[width=\linewidth]{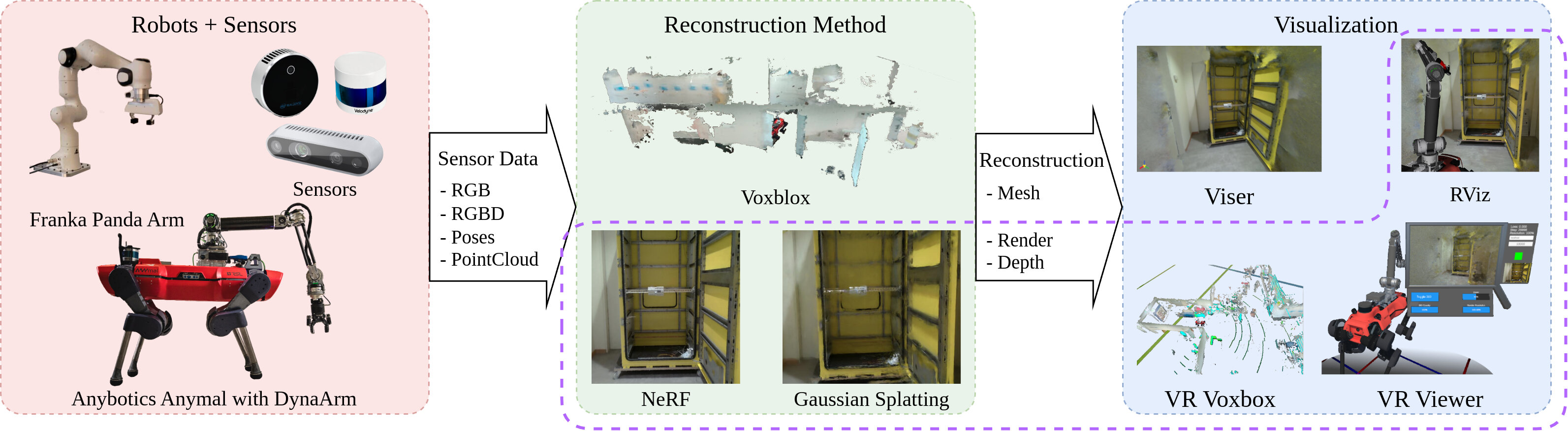}
    \caption{General teleoperation visualization pipeline divided into three sections: Robots, Reconstruction Methods, Visualization. Sensor and pose data flows from various robotic components (red) into the reconstruction method (green) to create a scene representation that is shown to the user in the visualizer (blue). Support for Radiance Field reconstructions such as Neural Radiance Fields (NeRFs) and Gaussian Splatting, as well as RViz and VR visualizers for these methods, are presented in this work. For baseline comparisons, a Voxblox mesh viewer was also ported to VR. These contributions are highlighted with a purple dashed line.}\vspace{-15px}
    \label{fig:pipeline}
\end{figure*}

The foundation of Neural Radiance Fields (NeRF)~\cite{nerf} has marked a significant leap in 3D reconstruction due to their ability to render novel photorealistic views from a sparse set of posed images. 
Recent advancements in NeRFs~\cite{2020posenc, mipnerf, instantngp} have significantly improved training and rendering time, making it an excellent candidate to improve the visualization and immersiveness of teleoperation systems.
However, NeRFs still struggle with computational efficiency and scene controllability~\cite{splat}, essential for teleoperation. 
In response to these challenges, another technique known as 3D Gaussian Splatting (3DGS)~\cite{splat} has emerged, offering an efficient approach to rendering. Unlike NeRFs, 3DGS utilizes an explicit Radiance Field representation, coupled with highly parallelized processing, to achieve efficient computation and rendering times. This is especially beneficial in complex settings, where precise manipulation of geometries and adaptation to lighting conditions are imperative.

With this in mind, this work aims to advance the state-of-the-art in robotic teleoperation by presenting a novel dynamic visualization pipeline that merges Radiance Fields with real-time robots. First and foremost, our system can train high-quality Radiance Fields from real-time robotic data in the form of a Radiance Field Node. The system is dynamic enough to handle data from any ROS-based robot, from a simple robotic arm to a highly mobile quadruped. Secondly, to keep up with the rapid advances in neural rendering, this system includes deep integration with NerfStudio~\cite{nerfstudio}, allowing for support for different methods, including Gaussian Splatting. Finally, the system provides a suite of visualization options allowing teleoperators to understand their environment through Radiance Fields. We offer an RViz-compatible Radiance Field Plugin system that integrates seamlessly with most existing ROS teleoperation systems, complete with multiple render modes, scene cropping, and depth-informed clipping. For more immersive use, a VR visualization suite is designed, allowing the user to stand in a virtual replica of the scene or send commands through a birds-eye-view system. Ultimately, this research opens up new possibilities for combining not just teleoperation with Radiance Fields but the use of online Radiance Fields for robotics in general.

\section{Related Works}

This section provides a brief overview of teleoperation and reconstruction methods, and recent work in Radiance Fields.

\subsubsection{Teleoperation and Reconstruction}
As robots are deployed to more complex uses such as industrial settings~\cite{magiclens} or caves and other natural environments~\cite{compslam}, there is an increased need for teleoperation. Here sensor data is streamed to an operator, either as direct camera feeds~\cite{xprize} or scene reconstructions~\cite{voxblox}. The state-of-the-art setups favor the high fidelity of direct camera systems~\cite{xprize}. However, they suffer from limited maneuverability due to fixed sensor feeds. Meanwhile, reconstruction-based teleoperation builds a scene representation that is easy to maneuver\cite{voxblox,fusion}, but often suffers from low fidelity or poor performance~\cite{voxfield}. Even under the best circumstances, these systems fail to represent volumetrics such as smoke and ignore view-dependent data such as specular effects. Radiance Fields are explored as a high-fidelity and maneuverable scene representation.

\subsubsection{Radiance Fields}
Originally used to render novel views with photorealistic quality, Radiance Fields have recently been adapted for large-scale 3D reconstructions. Neural Radiance Fields (NeRFs)~\cite{nerf} could capture small scenes by encoding geometry, color, and density into an MLP, allowing for view-dependent colors and volumetric rendering. Rapid advances allowed for high-frequency details~\cite{2020posenc}, unrestricted view construction~\cite{mipnerf}, and even sped up operation to near real-time on small scenes~\cite{instantngp}. 

As a faster alternative to NeRFs, Gaussian Splatting (3DGS)~\cite{splat} leverages a more explicit scene representation for real-time rendering. Representing the scene with 3D Gaussians can capture view-dependent color, volumetric elements, and unbounded scenes at higher speeds. However, it lacks the scene fidelity of NeRF, particularly with complex reflections with specular objects and sharp edges.

The recent explosion of novel NeRF methods has been aided by the creation of NerfStudio~\cite{nerfstudio}. NerfStudio provides a comprehensive system for building, training, and adapting Radiance Field methods. It includes methods such as Nerfacto (NeRF implementation) and Splatfacto (Gaussian Splatting implementation), as well as a variant of Instant-NGP, depth-supervised methods, and numerous others.

Until recently, the intersection of robotics and Radiance Fields was limited due to long training times and the difficulty in porting the systems to mobile hardware. Instead, Radiance Fields have mainly been used to perform localization with offline data~\cite{locnerf} or offline path planning~\cite{chen2024splatnav}. Online integration of NeRFs is available in NerfBridge~\cite{nerfbridge}, although lacks support for multiple cameras and newer methods. This work addresses these issues by enabling online multi-camera Radiance Field reconstruction, which easily supports new methods such as 3DGS.

\section{Methodology}

\begin{figure*}[t]
    \centering
    \includegraphics[width=\linewidth]{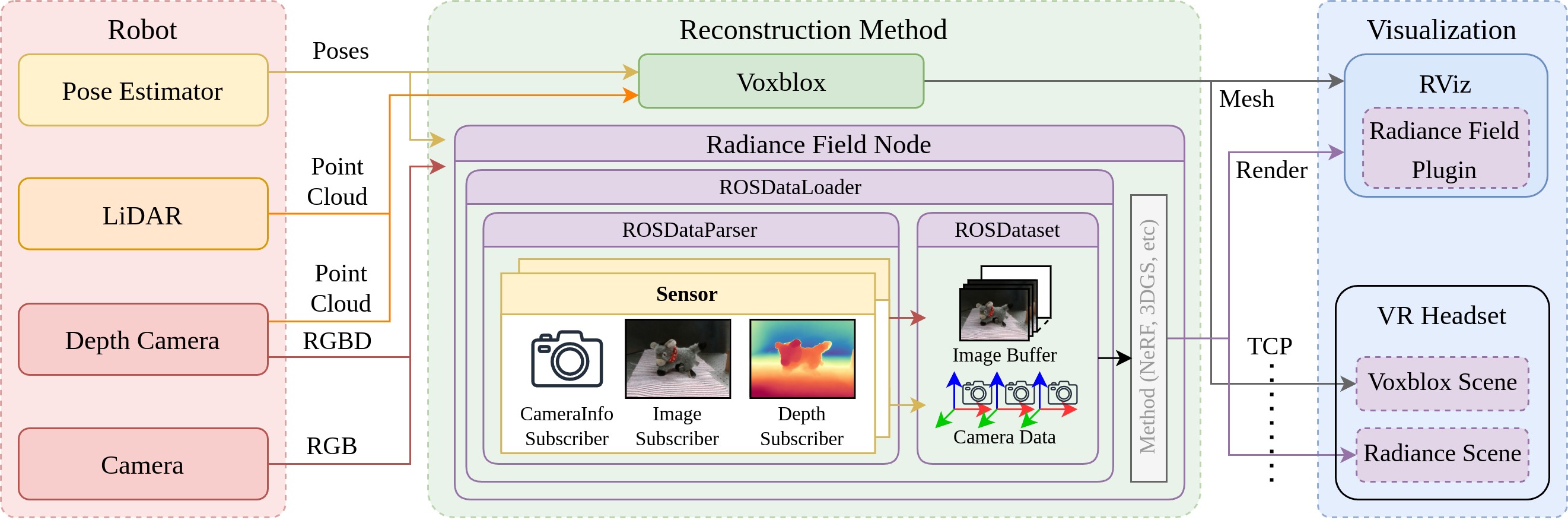}
    \caption{Component diagram showing the connections between the robotic systems (red and orange), the reconstruction methods (green), and the visualizers (blue). Data flows from the robotic systems into reconstruction method nodes where it is either merged into rendered views for Radiance Fields or a mesh for Voxblox. The Radiance Field Node is comprised of a custom DataLoader, DataParser, and Dataset, and uses a collection of Sensor objects to manage the ROS subscribers. The DataLoader and Dataset send data to any standard method during training, only returning images and cameras that have been updated. This data is then displayed directly in RViz or transmitted over TCP to a VR headset. }\vspace{-15px}
    \label{fig:overview}
\end{figure*}

The ideal teleoperation system would present the operator with a high-fidelity, maneuverable, and faithful representation of the scene~\cite{vr2}. This system should be robot-agnostic with minimal reconfiguration and be able to present environmental data--sensor streams or reconstructions--to best suit the task~\cite{vrteleop}. The extended 3D reconstruction-based pipeline, shown in~\cref{fig:pipeline}, involves taking sensor data from the robot over ROS and feeding it into a reconstruction method~\cite{voxblox,nerfstudio}. This method generates a representation of the environment relative to a fixed position which is presented to an operator via the visualizer. By constructing the representation relative to a fixed point, it can be displayed and aligned with additional data even as the robot moves through the space. Overall, we can simplify this pipeline into three components: the robot and its sensors, the reconstruction method, and the visualization system.

Here, we present a system that builds off existing teleoperation pipelines and extends the possible reconstruction methods to include Radiance Fields and a means of visualizing them both on screen and with a VR headset. This is highlighted in purple in~\cref{fig:pipeline}. The system was tested with multiple robots, including a simple static configuration, a mobile quadruped capable of exploring a larger environment, and a quadruped with an attached arm. This robotic data is sent to an existing mesh reconstruction as a baseline and a novel ROS node for Radiance Fields that supports NeRFs and 3DGS. These reconstruction methods are available in multiple visualizers, ranging from the 2D RViz window on the screen to 2.5D and 3D views in a VR headset shown in~\cref{fig:nerfvr}. Each component of this pipeline can be swapped out or reconfigured depending on the deployment and task.

\subsection{Robot}

The purpose of the robot in this system is to capture the environment to enable the user to understand the scene. The primary forms of data used in this pipeline are poses from ROS’ TF system, color images from onboard cameras, and RGBD images containing depth data and color information. To ensure the system is robot agnostic, these data sources can be easily configured for any deployment. In~\cref{fig:pipeline} and~\cref{fig:overview}, the incoming data sources are shown on the left inside the “Robot” box, with the red lines indicating image data, orange showing the point cloud data, and yellow showing the poses from the TF system.

The simplest robot setup would be a grounded robot, such as a robotic arm attached to a table. This robot has limited mobility, particularly when it comes to scanning a target object or a large-scale scene. However, it offers incredibly accurate poses, as the robot’s base frame is the fixed global frame. This gives reconstructions high pose confidence at the cost of scene scale and view angles.

A mobile robot is required to capture larger, more complex scenes. In this setup, the robot can move in the environment and capture virtually data from any angle. However, the relation between the robot and a fixed global frame is uncertain, requiring the mobile robot to localize within the environment. This can be accomplished with simultaneous localization and mapping (SLAM) systems such as CompSLAM~\cite{compslam}. The poses can drift during locomotion due to reliance on inaccurate onboard sensors. While the mobile base allows for greater coverage, the poses may not always be reliable.

One method to help ensure we can capture larger scenes with high pose accuracy and mobility is to affix a robotic arm to the mobile platform. In this setup, the robot can move around the scene to novel view angles and primarily collect data from the arm's motion while the base is static. As the robot base is static, the end-effector motion can be measured purely from the precise kinematic joint sensors. This avoids visual or LiDAR localization errors.

\subsection{Reconstruction Methods}

Once data is captured by the robot, it is passed into a reconstruction method. The purpose of this stage is to convert the numerous data streams into a single stream that an operator can use to control the robot. This involves creating a representation that is accurate both in terms of geometry and texture fidelity. This work compares two main approaches to this problem: mesh reconstruction via Voxblox~\cite{voxblox} and Radiance Field reconstruction via NerfStudio~\cite{nerfstudio}. These two ROS nodes are shown as green in the “Reconstruction Method” section of~\cref{fig:pipeline} and~\cref{fig:overview}. The Voxblox node generates an output mesh using incoming pose data and point clouds, while the NerfStudio node generates rendered Radiance Fields from poses and image data. As mesh reconstruction struggles to capture complex volumetric scenes, and trades-off between fast reconstruction and high fidelity, we propose using Radiance Fields for robotic teleoperation. Radiance fields seek to use lightweight machine learning algorithms in order to learn representations of a scene.

\textit{Radiance Field Node:} To maintain interoperability with new methods, the Radiance Field training node in the pipeline can be set to use nearly any method supported by NerfStudio~\cite{nerfstudio}. The number of custom components in the NerfStudio pipeline was minimized to ensure compatibility with any method so long as it doesn't require a custom Dataset (such as semantic models) or a custom DataLoader. This supported NerfStudio's 3DGS method: Splatfacto, with minimal overhead. At the start, this custom Dataset allocates a buffer of preconfigured size to hold all images for training. The node then subscribes to a set of topics for each camera to capture its images (RGB or RGBD), intrinsics, and poses. If all the cameras have depth topics, then a special buffer is allocated to store the depth images, allowing for integration with depth-supervised models.  Each incoming image is filtered to ensure it is not blurry and is a sufficiently different pose based on the mean distance between TF frames. If the cameras have different resolutions, all images are resized to the largest camera and scaled back down when collated into batches. For our experiments this meant storing image at 1440p with the 720p images being resized before training. The associated per-camera intrinsics and extrinsics are stored for each image based on the current pose data and the last CameraInfo message. If the image is already rectified and contains "rect" in the topic name, the distortion parameters of the associated camera are set to zero to avoid rectification. 

To ensure compatibility with additional methods, this package overrides the DataLoader with a custom ROS variant which skips the caching and collating normally performed, and instead uses the ROS Datasets functions to ensure only updated images are pulled for training. There is also a special evaluation DataLoader which can accept a list of image sequence IDs to ensure the same images are used across multiple runs. This is particularly helpful when running comparisons across prerecorded ROS bag data. Additionally, a custom DataParser is used to setup the ROS Dataset as well as ROS subscribers which are managed through a helper Sensor class. These Sensor objects manage the subscribers for image and CameraInfo messages, and are responsible for looking up the synchronized poses at each capture. The use of this helper class allows multiple cameras to be registered, each capturing data at a different rates. The Sensor objects will then update only the appropriate data entries, and forward the associated camera parameters to the rest of the system. The flow of data is shown in \cref{fig:overview} with all the custom components highlighted in purple and the Sensor helper class in yellow. Once everything is set up, the node captures several images to create the initial batch before switching to online training mode.

Once the node is in online training mode, it continues to receive new images up to the specified buffer size and begins an ActionServer, which allows any ROS process to request render data. A render request containing the view pose is sent to the server, and a rendered image, along with estimated scene depth is returned. These renders are then sent to one of the visualizers for the teleoperator to interact with. Additionally, the requests are associated with unique client IDs, allowing multiple visualizers to request different renders with a single Radiance Field node in the ROS network.
\subsection{Visualization}

The final stage of the pipeline is the visualizer. This program is intended to present an accurate representation of the world to the operator based on the data received from the reconstruction method. This is shown in blue in~\cref{fig:pipeline} and~\cref{fig:overview} , with our custom Radiance Field RViz plugin acting as the 2D visualization and the VR scenes as the 2.5D and 3D visualizers. For a baseline comparison, the Voxblox RViz plugin is used to show the mesh reconstruction.

\subsubsection{RViz}

\begin{figure}[ht]
    \centering
    \includegraphics[width=0.48\linewidth]{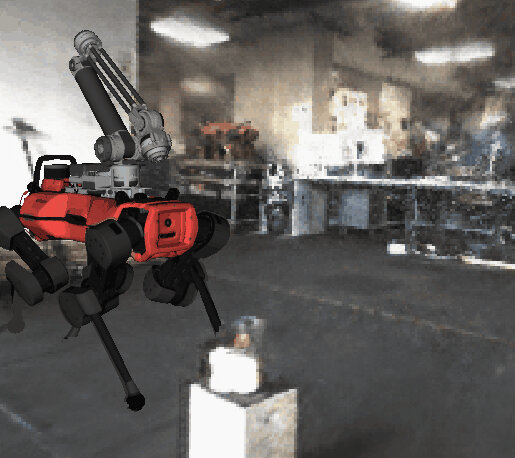}
    \includegraphics[width=0.48\linewidth]{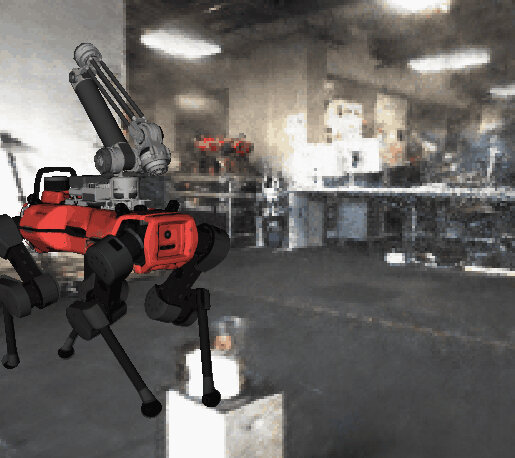}
    \caption{Sample of the RViz plugin occluding the robot based on scene depth (left) and having the robot rendered on top (right).}
    \label{fig:occlude-overlay}
\end{figure}

RViz is the de facto visualization suite for ROS, as it can visualize different sensor streams, such as point clouds, robot models, and images. Also, it contains tools to help users send commands directly through the system, such as pose goals. It offers support for 3rd-party plugins, allowing reconstruction methods such as Voxblox to implement custom viewers. This seamlessly integrates the new environmental data with existing robotic and teleoperation stacks. 

\textit{Radiance Field Plugin:} To integrate the online Radiance Field generation with ROS, a Radiance Field RViz plugin is presented. This plugin acts as a custom camera for the RViz OpenGL scene, sending render requests to the Radiance Field node whenever the camera is moved (dynamic mode) or a stream of renders based around a moving frame (continuous mode). Dynamic mode is best used when viewing the scene from a static perspective, such as an overhead view for navigation, or a close up for inspection. Continuous mode on the other hand is helpful for a moving base frame, such as third-person piloting, or when new incoming data needs to be observed. In either mode the speeds of the renders are often most dependant on resolution of the output image, which can be adjusted with a setting in the View panel. In order to provide rapid data for dynamic mode, the renders are sent first at 10\% of the final resolution, and then 50\% before sending over the full image. This allows the user to move around the scene with more rapid response from the rendered environment. 

Depth data from RViz and the Radiance Field are then used to either show the render realistically occluding the scene elements or as a cut-out, as shown in~\cref{fig:occlude-overlay}. 
This is achieved by transforming the real-world depth captured in the rendering into OpenGL's z-depth, a nonlinear scale ranging from 0 to 1. This transformed depth can then be matched with the z-depth used by RViz's Ogre engine (also based on OpenGL) for a specific camera perspective. As a result, all objects within RViz--such as robots, sensor visualizations, grids, and poses--are accurately occluded, providing a more realistic visualization. This aims to give the operator a better understanding of depth within the environment while also making it easier to locate RViz components in the scene. In large and indoor scenes, there is also a cut-out mode in which the rendered image is always displayed behind the RViz elements, making them easier to find.

\begin{figure}[t]
    \centering
    \includegraphics[width=0.48\linewidth]{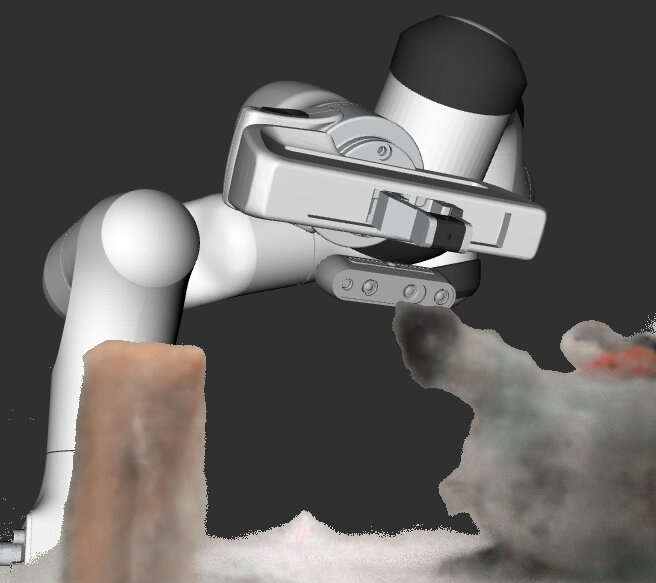}
    \includegraphics[width=0.48\linewidth]{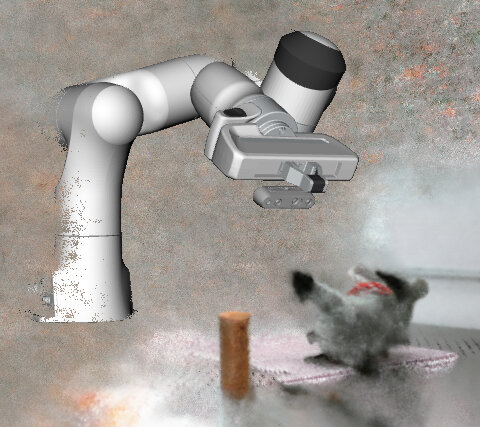}
    \caption{An axis-aligned bounding box can be used to crop the background (left) or remove walls (right) allowing for novel views and clearer operation.} \vspace{-15px}
    \label{fig:crop}
\end{figure}

Furthermore, as both 3DGS and NeRFs perform volumetric rendering, it can be difficult to self-localize due to occlusion or unexplored noise. To combat this, the plugin allows for the creation of an axis-aligned bounding box which crops the scene by limiting the ray integration. This is shown in~\cref{fig:crop} where the bounding box is used to remove the noisy background as well as the back wall that would obstruct the operator's view, visible in~\cref{fig:results}. These two renders are entirely novel views that the robot was unable to capture either due to the wall or limitations on the arm's mobility.

\subsubsection{VR}

\begin{figure}[h]
    \centering
    \includegraphics[width=0.51\linewidth]{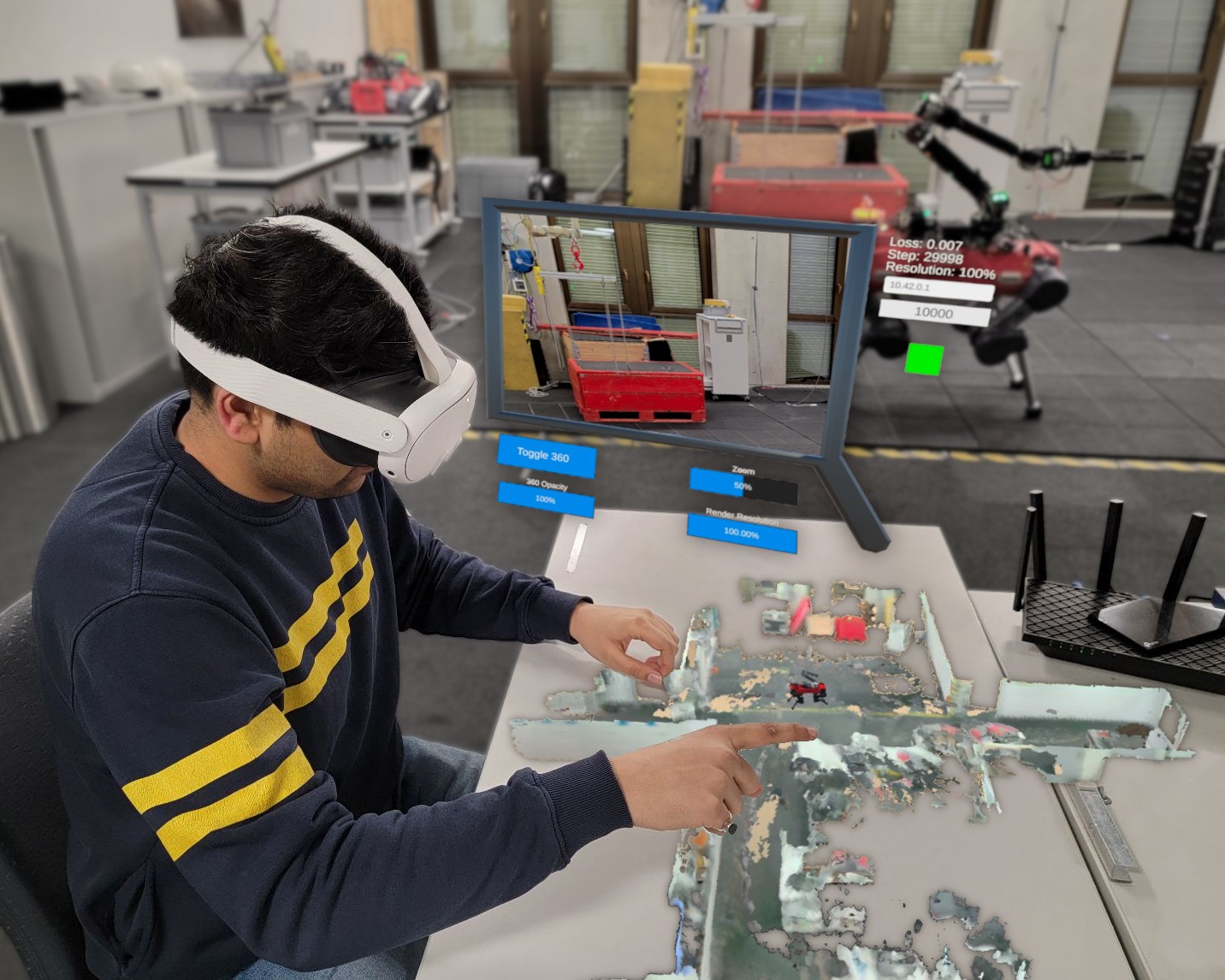}
    \includegraphics[width=0.41\linewidth]{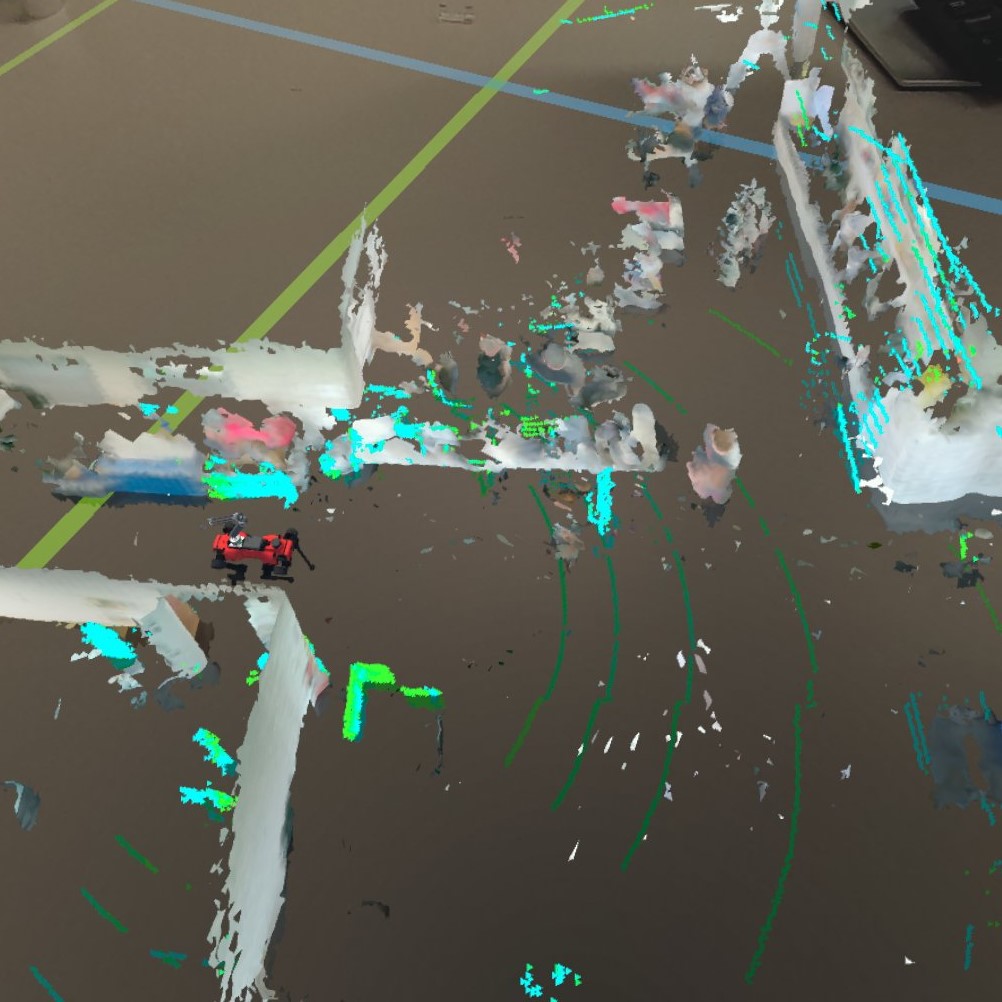}
    
    \caption{Teleoperator using VR system with Voxblox mesh on the table and Radiance Field Viewer on the left (left) and live LiDAR data overlayed on the mesh reconstruction.}
    \label{fig:voxvr}
\end{figure}

In testing, it was found that even with occlusion, it was difficult for an operator to get a sense of depth while looking at a 2D screen. To remedy this, the system was ported to a Meta Quest 3 VR headset running with a Unity scene to display the robotic data. This was following the prior works~\cite{vrteleop, vr2, magiclens} where VR was used to extend the capabilities of existing teleoperation systems for greater immersion and ease of use. 

\textit{VR Robot Data:} To create a baseline for comparison, a handful of RViz features, such as a TF synchronized robot model, pose publishing, and sensor visualizations (for LiDAR, RGBD point clouds, and images) were integrated into the VR headset as shown in \cref{fig:voxvr}. The user is able to send pose goals via the headset and hand interaction, enabling them to send robot commands directly from the VR interface. This system was extended with a mobile GPU-based mesh generation for real-time streaming of the Voxblox reconstruction. This provides the user with the same sensor information they can access in RViz, but displayed in an immersive 3D setting. In order to remain lightweight, the VR scene is optimized to run natively on the headset, with only a TCP connection to ROS handling message traffic. This allows the headset to connect directly to the ROS network of the robot and begin interaction with out the need for a base station PC or any headset tethers.

\begin{figure}[ht]
    \centering
    \includegraphics[width=0.48\linewidth]{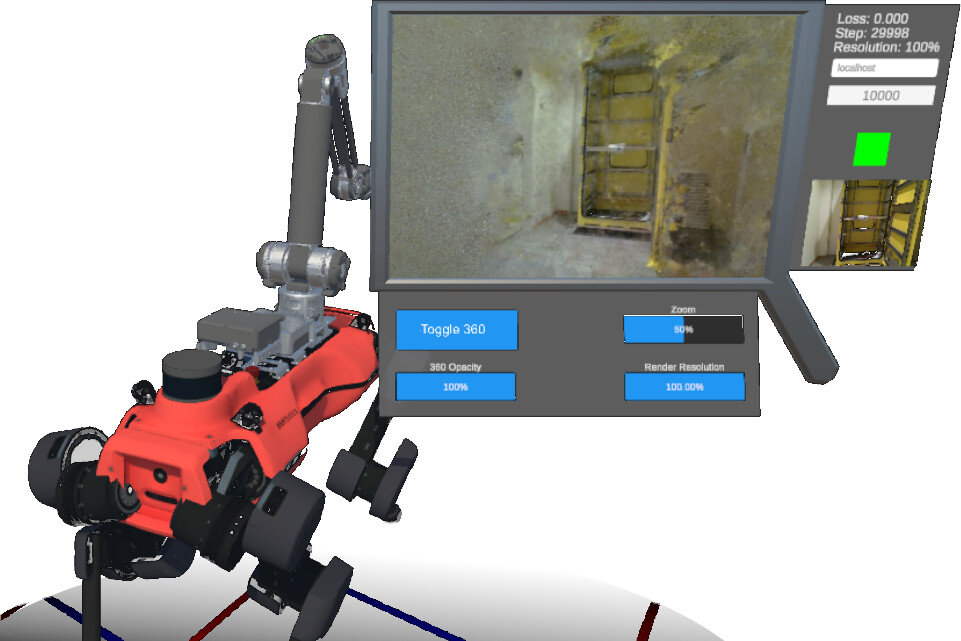}
    \includegraphics[width=0.48\linewidth]{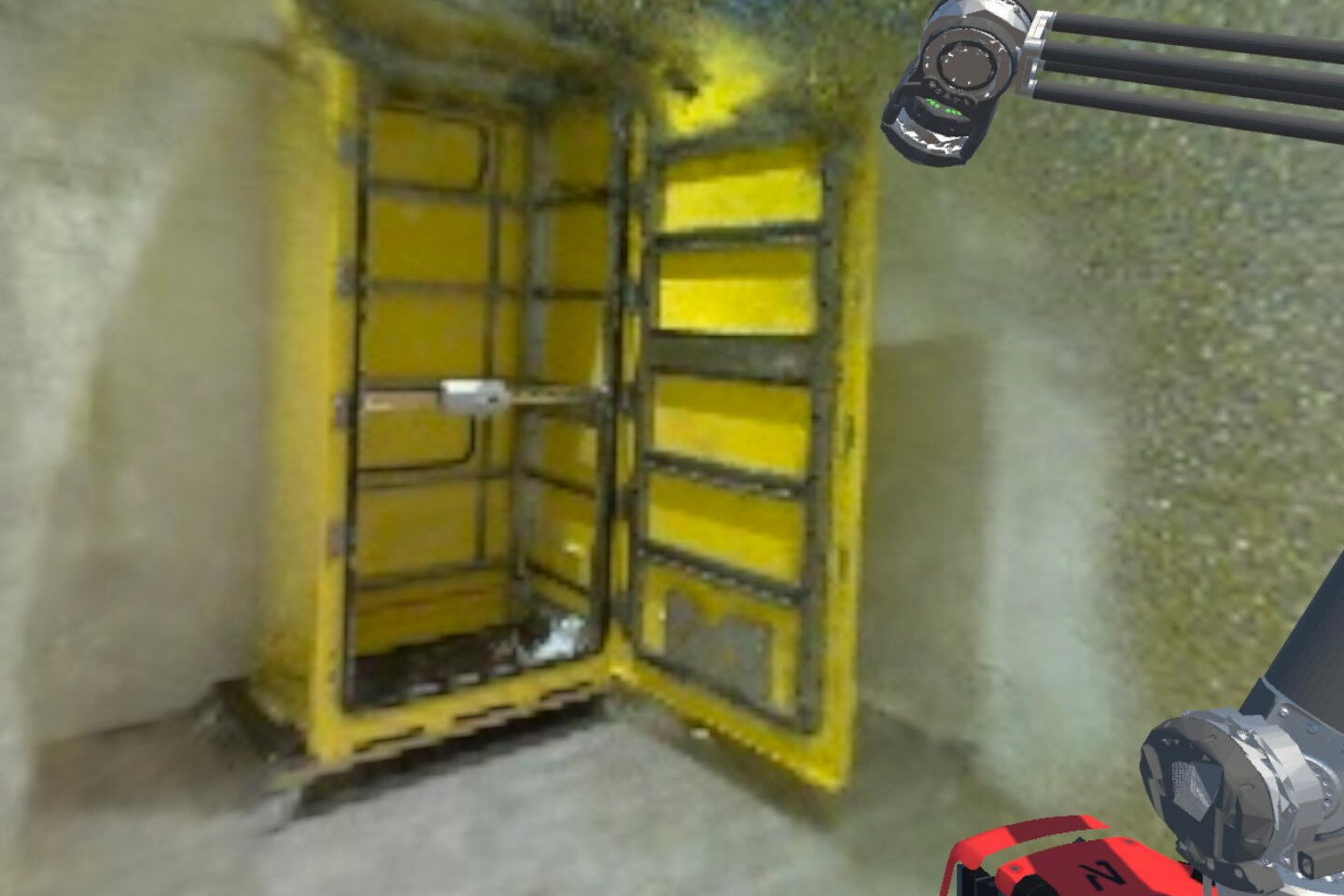}
    
    \caption{Radiance Field VR scene with 2.5D handheld parallax viewer (left) and fully immersive 360$^\circ$ (right).}\vspace{-10px}
    \label{fig:nerfvr}
\end{figure}

\textit{VR Radiance Fields:} To use the Radiance Field renders, there is a 2.5D handheld viewer, which provides the user a view into the rendered world. This viewer uses the depth data to create a parallax effect allowing the VR headset to provide a greater sense of depth while looking at the renders. Similar to the RViz camera, the viewer can request new renders whenever the user moves the head and it is synchronized to the position of the robot data in the scene. Requests to the ActionServer are directly routed through the TCP connection, and the rendered results are streamed straight to a GPU shader for parallax rendering. For further immersion, there is also a 360$^\circ$ spherical render which gives the user a sense of scale as if they are physically present in the environment. The viewer also offers controls such as image zoom, resolution, and a live camera feed from the robot. Figure~\ref{fig:nerfvr} shows the handheld Radiance Field viewer during deployment and the view from the spherical render mode. Similar to RViz plugin, the depth data from the Radiance Fields is also used to provide dynamic scene occlusion, allowing closer 3D objects, such as robots or point clouds, to be rendered in front of the Radiance Field renders. 

\section{Experiments}

As there are three major components to this pipeline, three sets of experiments were conducted. First, data was captured on a static arm, mobile base, and mobile arm to ensure robot-agnostic operation and compare reconstructions. Each dataset was processed with Voxblox, NeRF, and 3DGS, and photometric quality was evaluated. Second, to test online operations, reconstruction and rendering times were measured. Lastly, to validate the viewing experience, a user study was conducted to compare the RViz mesh and NeRF renders vs their VR counterparts.

\subsection{Datasets}
\subsubsection{Static Arm}
A static Franka Panda arm was used with an Intel Realsense 435i RGBD camera. The arm was fixed to a metal table and bounded by walls on two sides, reducing the target scanning area to 20cm x 20cm. The third side of this area was unexplorable as it is where the arm was mounted, limiting the range of scans to front-facing and only 90$^\circ$ of azimuth. Three targets were captured: a stuffed donkey toy with fine fur, a fiducial cube, and a wooden block. The fur was used to test the high-frequency details, the fiducials checked texture accuracy, and the block was used to confirm 3D geometric reconstruction.

\subsubsection{Mobile Base}
An Anybotics Anymal was used to walk around a large lab environment. The scale of the scene was roughly 15m x 10m and was captured with front and rear-facing 1440p RGB cameras. Localization was performed using CompSLAM~\cite{compslam} and centered around a wooden pedestal which the robot orbited. Atop the pedestal was a glass bowl, testing each system's transparent and specular reconstruction abilities.

\subsubsection{Mobile Arm}

A DynaArm was attached to the Anybotics Anymal with an additional Intel Realsense L515 mounted at the wrist. The target of capture was a yellow switchboard cabinet with a metal bar across the middle holding a box with a screen and switch. This was roughly 1m x 1m and captured from the front, with Anymal scanning 2m away with the arm, before moving in and capturing the inside of the box. As the arm moved into the box, there was a slight change in lighting condition, testing how well the systems adapt to dynamic colors.

\subsection{Quality}

\begin{figure}[ht]
\centering

\renewcommand{\tabcolsep}{0pt}
\begin{tabular}{cccc}
     \includegraphics[width=0.25\linewidth]{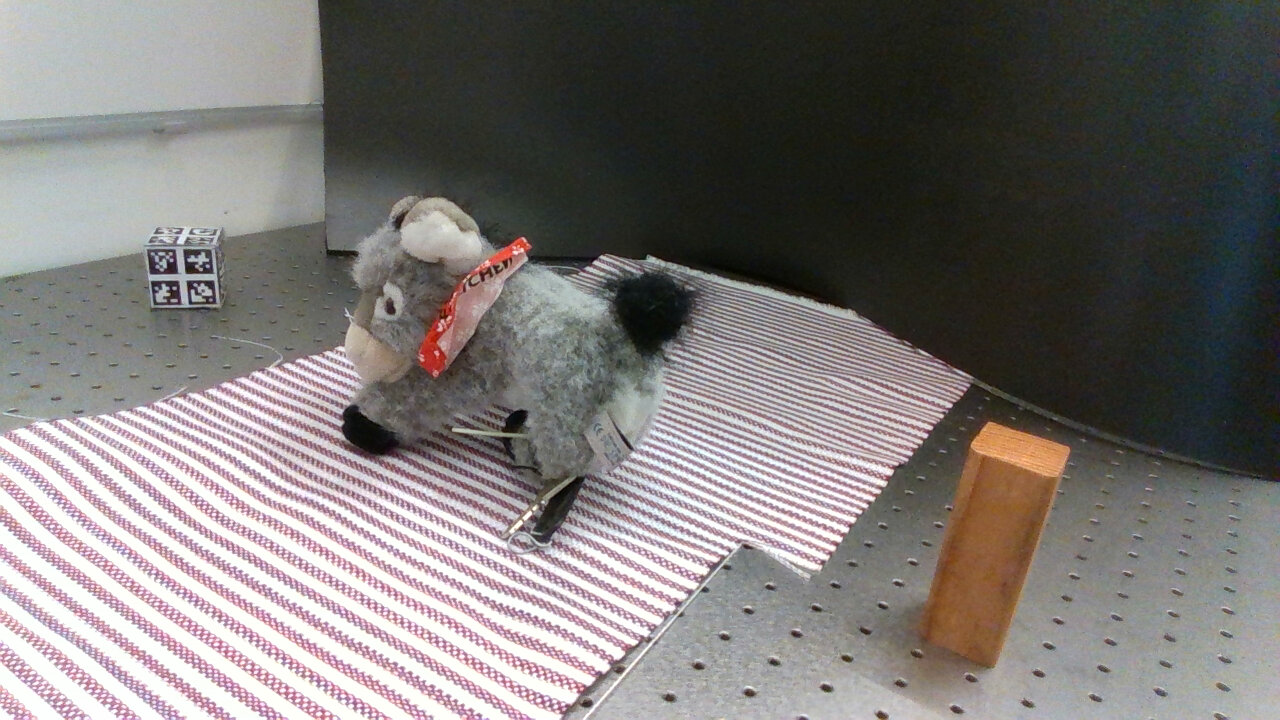}     &  \includegraphics[width=0.25\linewidth]{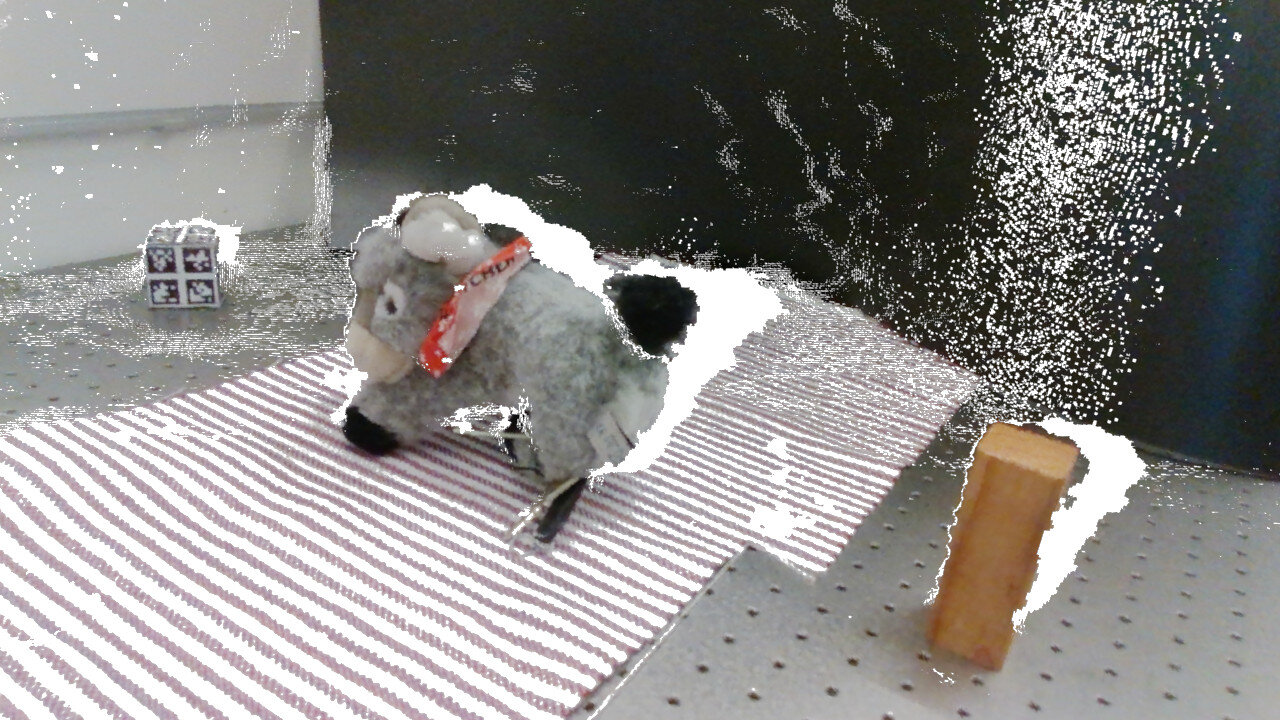}    & \includegraphics[width=0.25\linewidth]{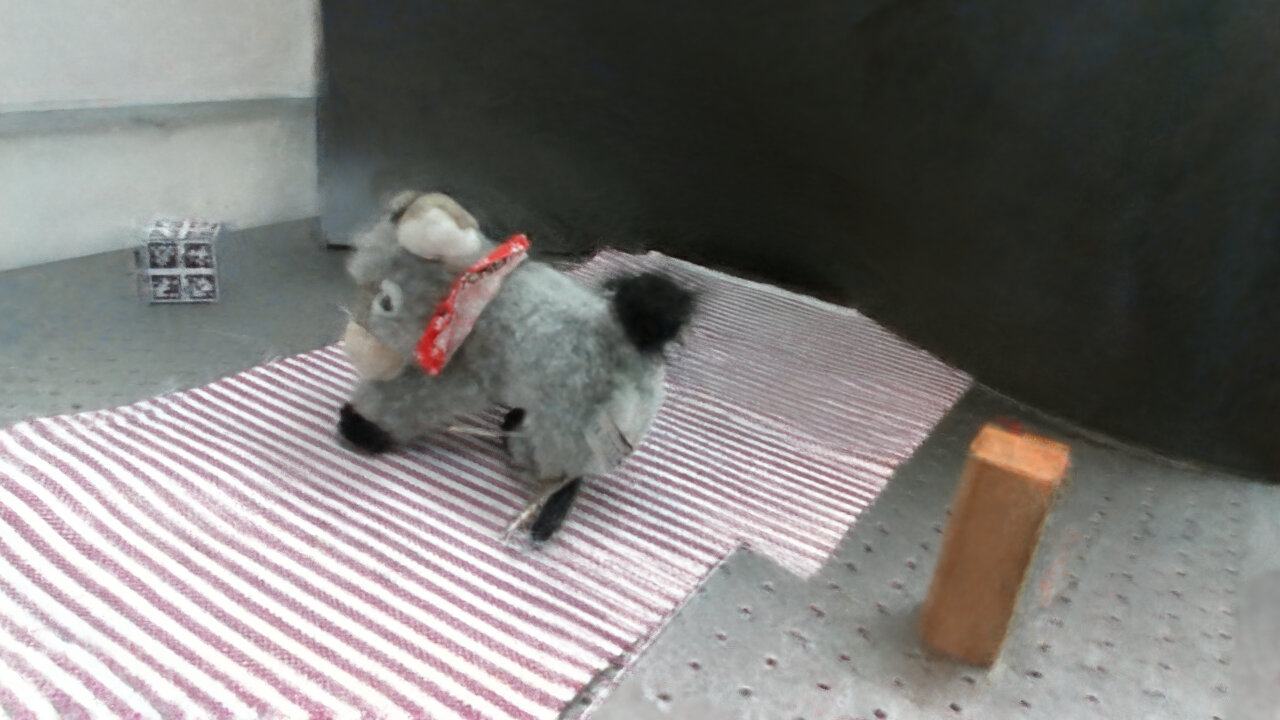}   &  \includegraphics[width=0.25\linewidth]{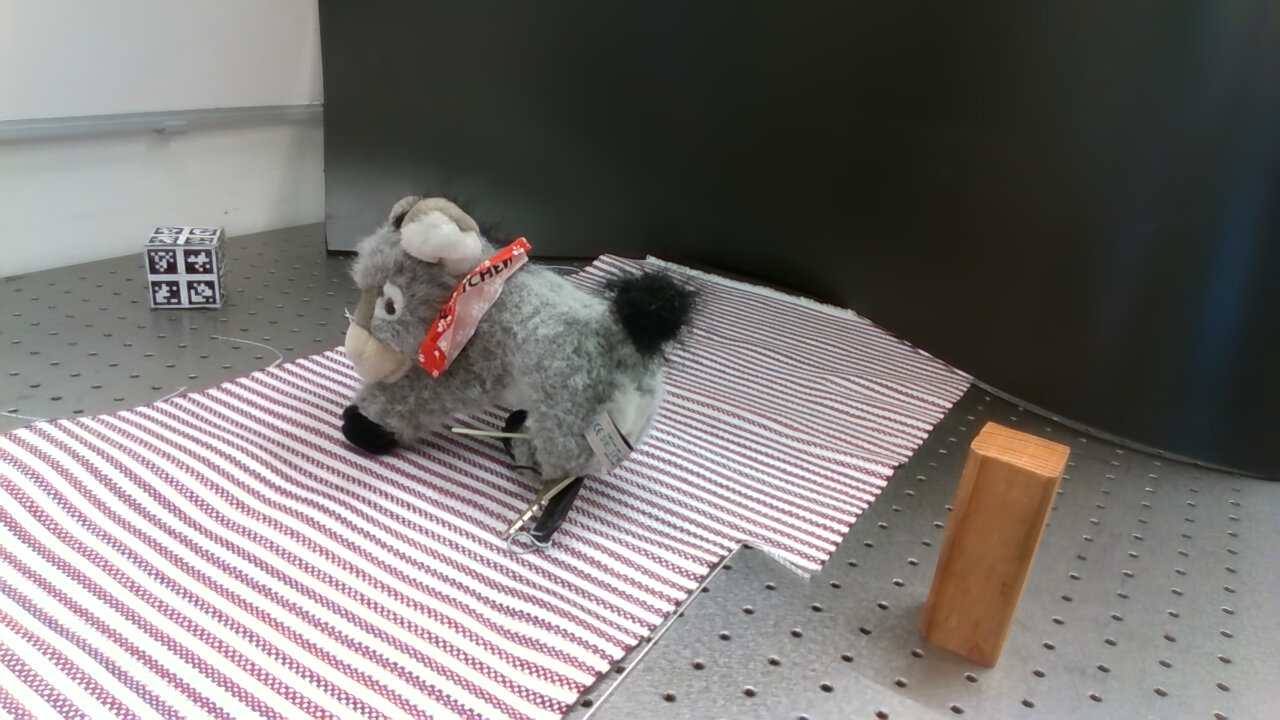}  \\
     \includegraphics[width=0.25\linewidth]{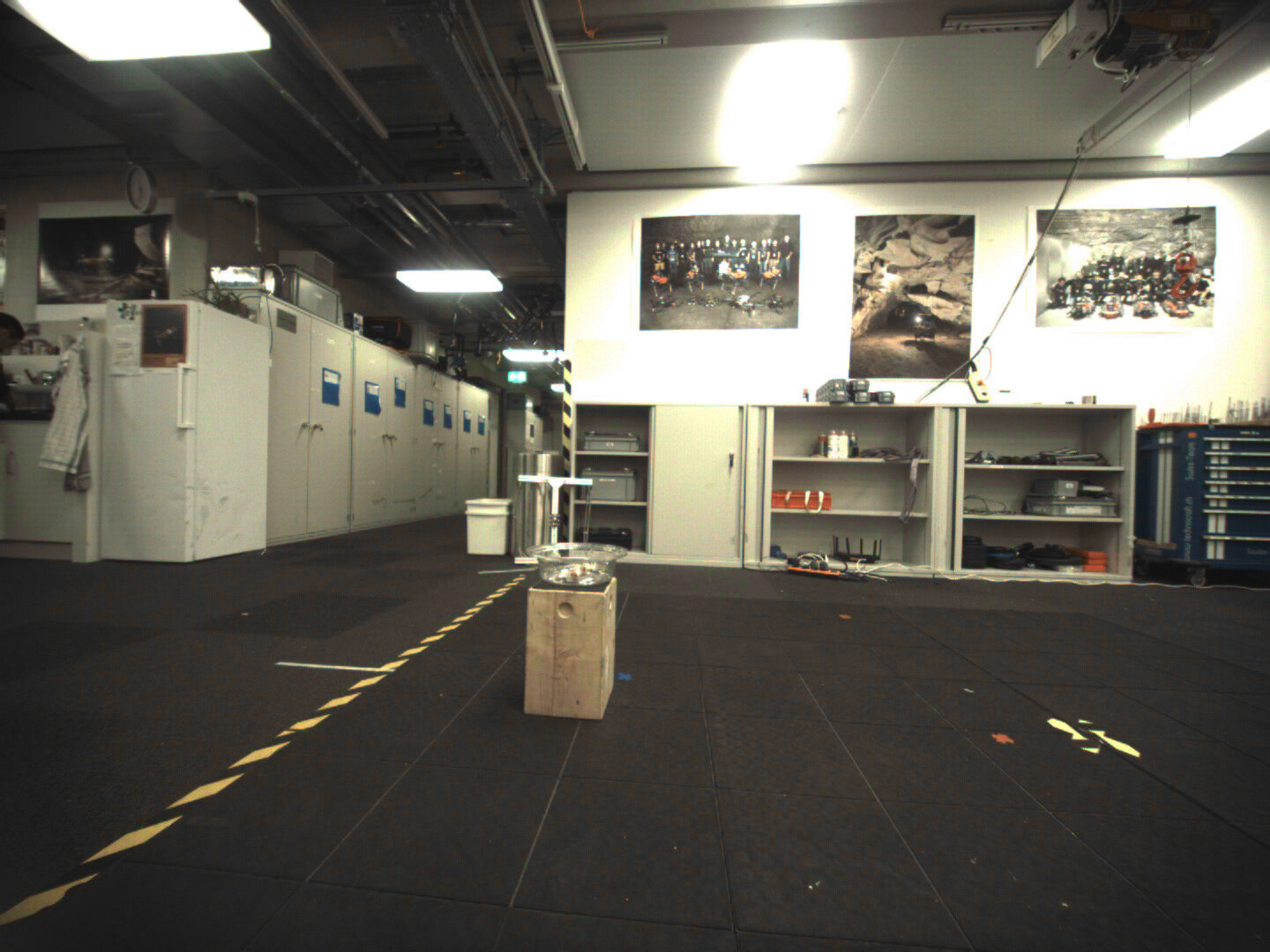}    &  \includegraphics[width=0.25\linewidth]{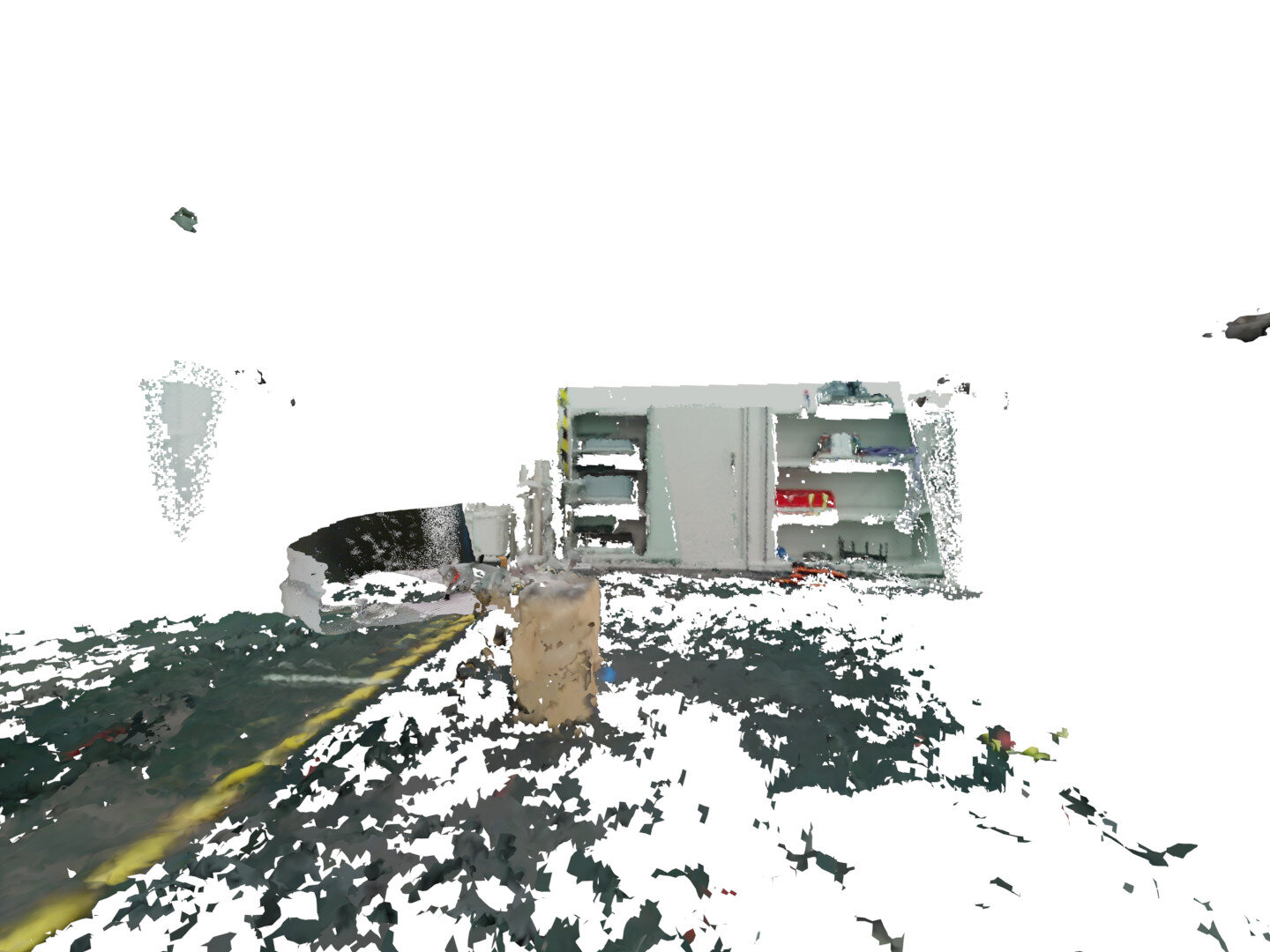}   &  \includegraphics[width=0.25\linewidth]{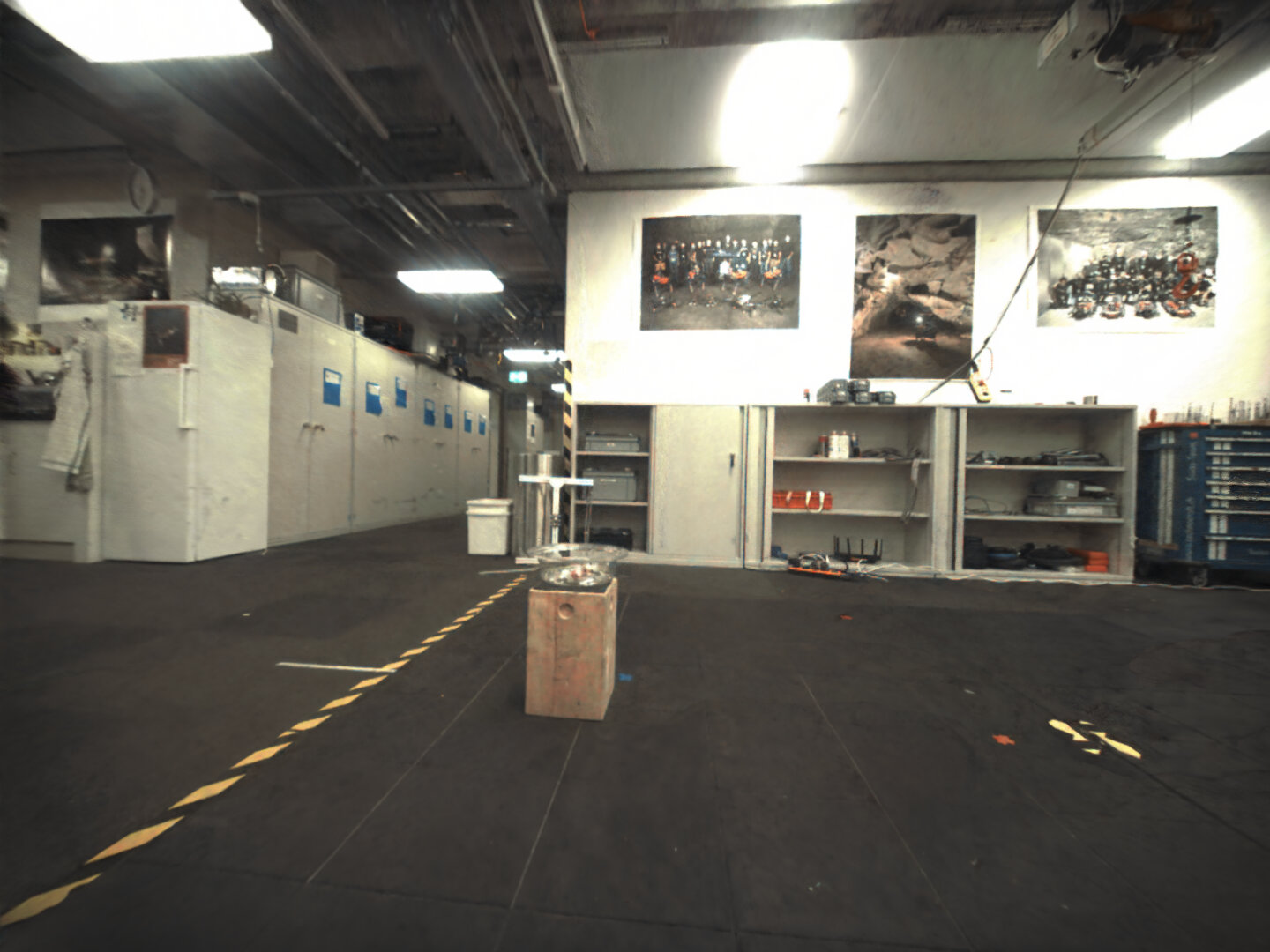}  &  \includegraphics[width=0.25\linewidth]{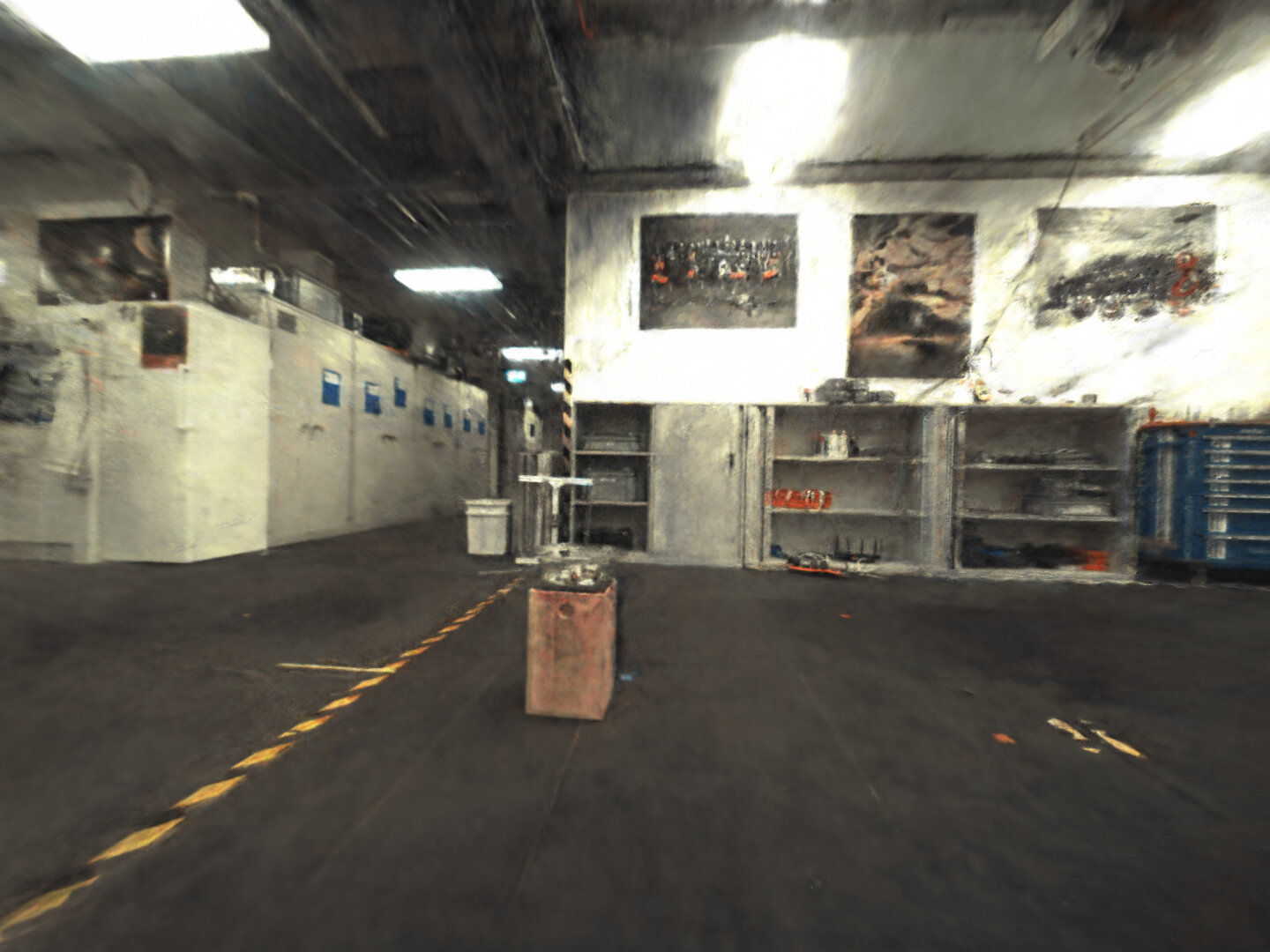} \\
     \includegraphics[width=0.25\linewidth]{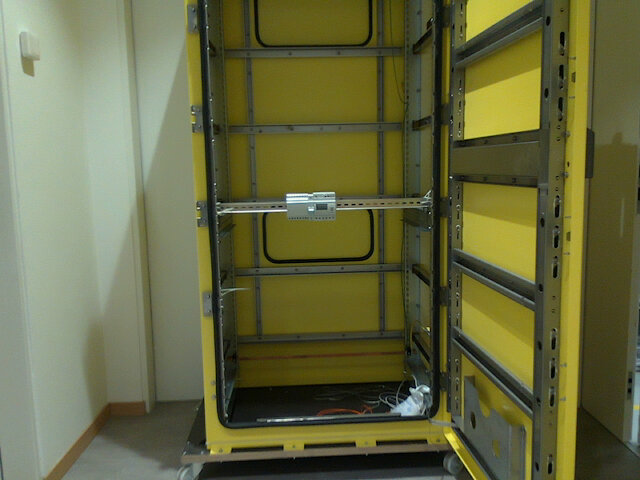}      &  \includegraphics[width=0.25\linewidth]{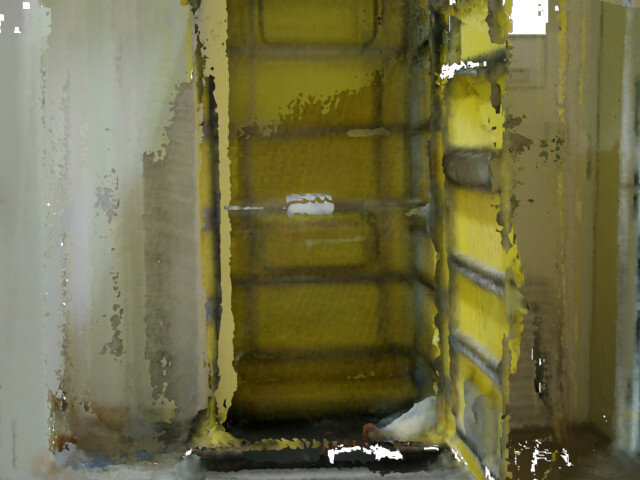}     & \includegraphics[width=0.25\linewidth]{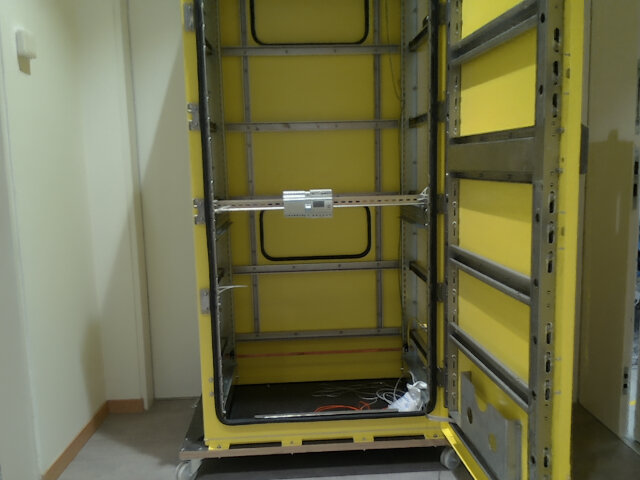}    &  \includegraphics[width=0.25\linewidth]{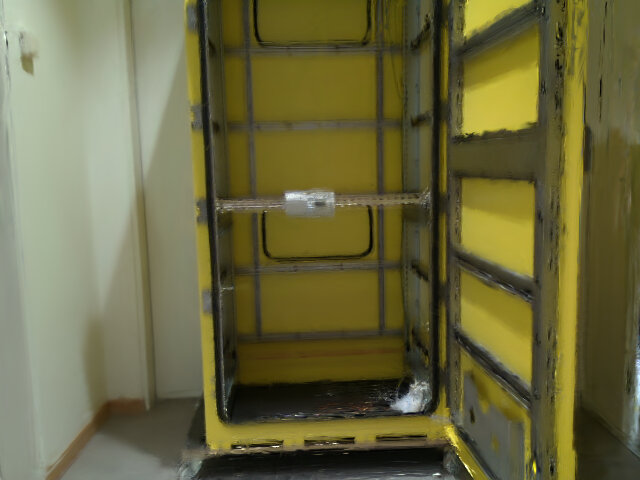}   \\
Ground Truth & Voxblox & NeRF & 3DGS
\end{tabular}

\caption{Comparisons of the static arm (left), mobile base (middle), and mobile arm (right). The top row shows the ground truth images, the second row shows the Voxblox reconstructions, the third row shows the NeRF reconstruction and the bottom row shows the 3DGS reconstruction. NeRF produced the highest quality results for both the mobile datasets, while 3DGS created a near-perfect image for the static arm dataset.}\vspace{-5px}
\label{fig:results}

\end{figure}

\begin{table}[h]
\centering
\caption{Photometric Comparisons of Different Robot Configurations and Reconstruction Methods }
\label{table:robots}
\begin{tabular}{r l c c c}
\toprule
Dataset & Method & PSNR $\uparrow$ & SSIM $\uparrow$   & LPIPS $\downarrow$  \\ \midrule \midrule
\multirow{3}{*}{\begin{tabular}[c]{@{}r@{}}Static Arm\end{tabular}}   & Voxblox & 15.42 & 0.4969 & 0.5507 \\
        & NeRF   & 18.07          & 0.5309          & 0.4917          \\
        & 3DGS   & \textbf{34.12} & \textbf{0.9288} & \textbf{0.2298} \\ \midrule
\multirow{3}{*}{\begin{tabular}[c]{@{}r@{}}Mobile Base\end{tabular}}       & Voxblox & 6.85  & 0.4091 & 0.8051 \\
        & NeRF   & \textbf{25.55} & \textbf{0.8718} & \textbf{0.1988} \\
        & 3DGS   & 20.18          & 0.7853          & 0.3953          \\ \midrule
\multirow{3}{*}{\begin{tabular}[c]{@{}r@{}}Mobile Arm\end{tabular}} & Voxblox & 16.94 & 0.514  & 0.5295 \\
        & NeRF   & 22.16          & 0.6473          & \textbf{0.1478} \\
        & 3DGS   & \textbf{24.59} & \textbf{0.7414} & 0.2126          \\ \bottomrule
\end{tabular}\vspace{-5px}
\end{table}

In order to ensure the platform is robot agnostic, three different deployments were run. All three setups were trained with the same parameters using Voxblox, Nerfacto, and Splatfacto, with Nerfacto and Splatfacto being the default NeRF and 3DGS methods from NerfStudio. Each was evaluated based on their peak-signal-to-noise ratio (PSNR), structural similarity index measure (SSIM), and learned perceptual image patch similarity (LPIPS)~\cite{lpips}. PSNR is used to measure the artifacts in the scene, SSIM measures the similarity for features such as lighting and contrast, and LPIPS measures the network activation of patches, approximating what humans would say is similar. The ground truth picture, baseline Voxblox reconstruction, NeRF, and 3DGS are shown in~\cref{fig:results}, while the photometric comparisons are shown in~\cref{table:robots}. The best-resulting method for each dataset is shown in bold.

\subsubsection{Static Arm}

As the Panda arm data had near-perfect poses, it was able to produce results of very high texture fidelity. High enough that in~\cref{fig:results}, the fiducial cube's markers are clear enough to scan. The lack of mobility produced limited views on the periphery of the scene, resulting in lower-quality NeRF reconstructions. However, the small scene size meant incredibly high-quality 3DGS reconstructions. This is because the scene was well initialized, with Gaussians densely covering the entire space with little excess. Across all metrics--PSNR, SSIM, and LPIPS--the 3DGS reconstruction performed twice as well as the NeRF. While the captured areas are decently represented, Voxblox created an incomplete mesh, scoring the lowest in all metrics. These results are shown in~\cref{table:robots}.

\subsubsection{Mobile Base}

Radiance Fields, particularly NeRFs, struggle with ray stability far from the scene center, resulting in noise. Even Voxblox struggled to capture distant data, producing a mostly incomplete reconstruction. During training, the NeRF method was able to smooth out the noise via pose optimization, resulting in the best results. Additionally, the target glass bowl was entirely missing from the Voxblox reconstruction, while it was clearly captured in the NeRF and roughly captured with 3DGS.

\subsubsection{Mobile Arm}

The switchboard cabinet produced the highest quality results for LPIPS on the NeRF and 3DGS reconstruction, representing the highest human-perceived quality. Voxblox managed to recreate most of the scene, although failed to capture the support beam and part of the door. Additionally, as the lighting changed when moving inside the cabinet, the mesh was not uniformly colored. This was captured in the Radiance Fields through their view-dependent coloring. As shown in~\cref{table:robots}, 3DGS outperformed NeRF on PSNR and SSIM, likely due to its better color matching with spherical harmonics. 

\subsection{Performance}

\begin{table}[]
\centering
\caption{Time Comparisons Between Different Methods}
\label{table:time}
\begin{tabular}{r c c c c }
\toprule
Method & \begin{tabular}[c]{@{}c@{}}Per Iteration\\ Time {[}ms{]}\end{tabular} & \begin{tabular}[c]{@{}c@{}}PSNR of\\ 16.94~dB {[}s{]}\end{tabular} & \begin{tabular}[c]{@{}c@{}}Render \\ Time {[}ms{]}\end{tabular} & \begin{tabular}[c]{@{}c@{}}Render \\ Time {[}FPS{]}\end{tabular} \\  \midrule \midrule
Voxblox & 1205.903          & 165.470           & --              & -- \\ 
NeRF    & 35.644            & 7.027             & 1020.131        & 0.980 \\
3DGS    & \textbf{34.651}   & \textbf{6.996}    & \textbf{6.6257} & \textbf{151} \\ \bottomrule

\end{tabular}\vspace{-15px}
\end{table}

Online robotic teleoperation requires fast reconstruction and visualization. To that end, each method's reconstruction and render times were compared on an RTX 4090. In order to ensure repeatability across runs and comparisons, the data was first recorded rosbags and the sequence ID for images was stored to a file. This was then used to create the Datasets during execution, ensuring the same images at the same timestamps were used in each run. First, the per iteration time was measured. For Voxblox, this is the time to integrate new batches with the mesh, while for the Radiance Fields, this was the training time for one iteration. Voxblox was run with 1cm voxels, and took 1.2 seconds to integrate each batch. For NeRFs and 3DGS, the iteration time was 35.644ms and 34.651ms, respectively. Additionally, for a fair comparison, the radiance methods were trained until they matched the final Voxblox PSNR of 16.94dB. For NeRF, this took 7.027s and 3DGS took 6.996s. Not only did the Radiance Field methods produce higher quality results, but they achieved this almost 20x faster. A summary of these results is shown in~\cref{table:time}.

\begin{figure}[h]
    \centering
    \includegraphics[width=\linewidth]{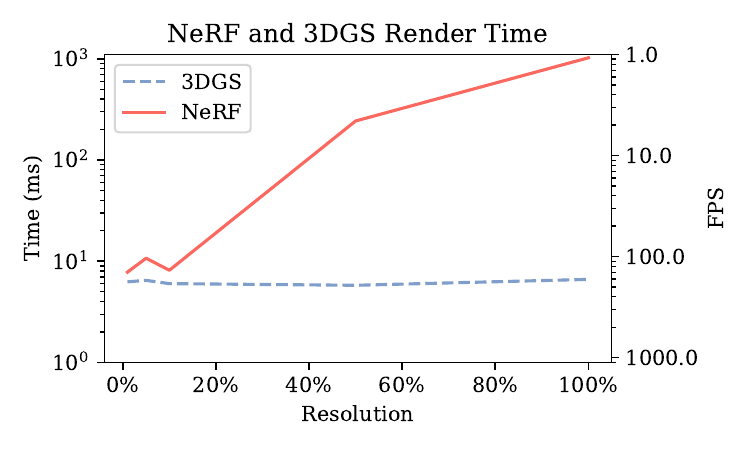}
    \caption{Render times for NeRF and 3DGS for a 1024x1024 image at different resolution percentages, showing an exponential increase in render time for NeRFs with a nearly constant time for 3DGS. The data is displayed at log scale due to the rapid increase in NeRF render times.
}
    \label{fig:times}
\end{figure}

Second, the render time was considered. As Voxblox displays a continually updating mesh, its render is limited to 30 frames per second by the viewer. For NeRF and 3DGS, a full 1024x1024 image was rendered, taking 1020.13ms (0.98FPS) for NeRF and only 6.63ms (151FPS) for 3DGS. When comparing the rendering times for different resolutions, 3DGS remains fairly constant at around 6.23ms, while NeRF renders take exponentially longer for larger resolutions. This relation is shown in~\cref{fig:times}. For use cases requiring rapid rendering, explicit representations such as mesh and 3DGS are preferred as they scale well for larger resolutions and screens.

\subsection{User Study}
\begin{figure*}[h]
    \centering
    \includegraphics[width=\linewidth]{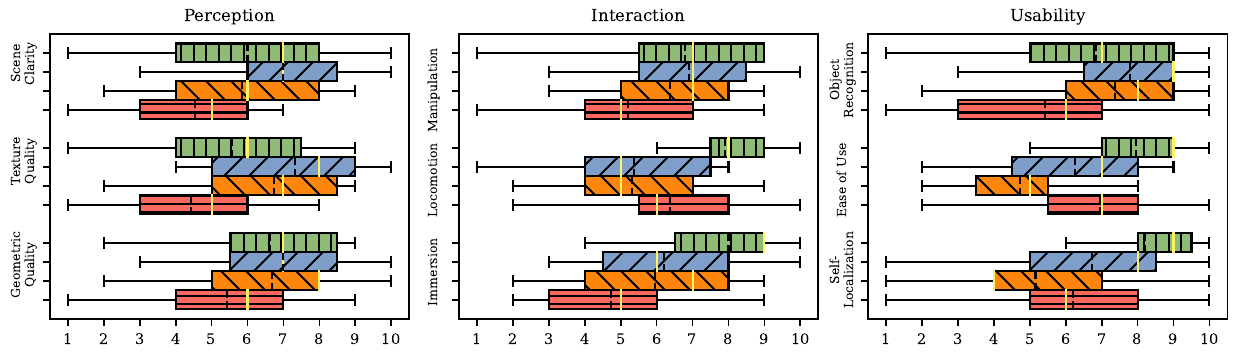}
    \caption{Results of the user study using the mobile arm dataset and four viewing modes: VR Voxblox(
    \begin{tikzpicture}
        \fill[preaction={fill=cgreen}][pattern=vertical lines] (0,0) rectangle (.5,.2);
    \end{tikzpicture}
    ), VR NeRF( 
    \begin{tikzpicture}
        \fill[preaction={fill=cblue}][pattern=north east lines] (0,0) rectangle (.5,.2);
    \end{tikzpicture} 
    ), RViz NeRF(
    \begin{tikzpicture}
        \fill[preaction={fill=corange}][pattern=north west lines] (0,0) rectangle (.5,.2);
    \end{tikzpicture} 
    ), RViz Voxblox( 
    \begin{tikzpicture}
        \fill[preaction={fill=cred}][pattern=horizontal lines, line width=6mm] (0,0) rectangle (.5,.2);
    \end{tikzpicture} ). The means are shown with a black dotted line, while the medians are the yellow solid line. In all cases, the VR systems were preferred to their 2D counterparts, while NeRFs were preferred for perception and manipulation and Voxblox for locomotion and usability.} 
    \label{fig:user}
\end{figure*}

A user study was conducted with 20 participants to compare the different visualization systems. The participants were chosen from a group familiar with robotic systems and who regularly use RViz. Their ages were between 22 and 32, with a mean age of 26, there were 5 female and 15 male participants, and only 5 had used VR systems before. They evaluated the mobile arm dataset in~\cref{fig:results} and compared the 2D RViz Voxblox and NeRF scene with VR counterparts, with the results shown in~\cref{fig:user}. This dataset was chosen as it had the highest photometric scores,~\cref{table:robots}, and therefore the highest quality. To avoid confusion, as both the Radiance Fields, NeRF and 3DGS, are presented via the same user interface, only one method was presented to the users. NeRF was chosen for this comparison as it outscored 3DGS in LPIPS, which best approximates human-perceived quality~\cite{lpips}. At the resolutions tested, the difference in render time between NeRF and 3DGS was imperceptible. 

Starting with perception, the NeRF was preferred to the Voxblox reconstruction, with the VR NeRF scoring slightly higher than the RViz variant. This confirms the results of the photometric comparison in~\cref{table:robots}, that Radiance Fields produce higher quality results than a mesh. Interestingly, the VR systems consistently increased the perceived quality, despite showing the same data as RViz. This is due to the immersion of the VR system, with the added depth helping geometric perception and the head-mounted optics helping boost texture quality. Additionally, the optics of the VR headset allow lower resolution images to be perceived as larger and higher resolution, decreasing the render latency for the system.

In terms of teleoperation tasks, the 3D mesh of Voxblox VR and RViz was preferred for locomotion tasks as the users could easily see where the robot was in relation to the environment. However, for manipulation, RViz NeRF was preferable to Voxblox, while the VR systems were roughly the same. Perception and ability to read fine details are paramount in these tasks, favoring the NeRF system, with many VR users saying they would prefer a 3D NeRF for the ideal setup. This can be addressed in the future by leveraging 3DGS' explicit representation to directly show the Gaussians in 3D without the need for the 2.5D viewer.

Finally, the VR systems were favored for usability over their RViz counterparts, despite this being the first time many users tried VR. The Voxblox systems were easier to maneuver and explore, as the direct 3D representation was intuitive. The RViz NeRF overlay and handheld viewer took time to get used to, but was the better platform for identifying objects within the scene.

\section{Conclusion}

In conclusion, this paper presents significant advancements in the field of robotic teleoperation through a comprehensive Radiance Field visualization pipeline integrating multiple cameras, dynamic support for new reconstruction methods, and VR integration. The experiments conducted demonstrate the effectiveness and versatility of the proposed system. 

Firstly, demonstrating the system's adaptability for different robotic deployments, from static arms to mobile bases with multiple cameras, in each setup, the Radiance Fields produced higher quality reconstructions compared to the mesh. Secondly, the greatly increased render speed of 3DGS makes it comparable to mesh rendering for online use while only requiring less time to hit the same quality. Lastly, the user study comparing visualization systems underscores a user preference for the quality of Radiance Fields over traditional meshes. For manipulation tasks, these renders were even preferred to the traditional system. Across the board, the VR systems provided a better experience, suggesting that a VR version of explicitly 3D Radiance Fields, such as direct 3DGS in VR, would be the optimal teleoperation setup.

Overall, this work contributes to advancing state-of-the-art robotic teleoperation by providing a robust pipeline adaptable to different deployments, offering high-fidelity reconstruction capabilities, and leveraging immersive VR environments for enhanced user experience. The findings suggest promising avenues for future research in the intersection of robotics and Radiance Fields technologies.

\addtolength{\textheight}{-12cm}   %

\bibliographystyle{bibliography/IEEEtran}
\bibliography{bibliography/IEEEabrv, bibliography/references}

\end{document}